\documentclass[11pt]{article}
\usepackage{amsmath}
\usepackage{times}
\usepackage{amsfonts,amssymb,float, mathtools,multirow, amsthm}
\usepackage{amsmath}

\DeclareMathOperator*{\argmin}{arg\,min}
\usepackage{bm,xcolor}
\newcommand{\bs}[1] {\bm{#1}}
\usepackage{natbib}
\usepackage{xr}
\usepackage{enumitem}
\usepackage{graphicx,url}
\usepackage[noblocks]{authblk}

\DeclareMathOperator*{\E}{\rm E}
\usepackage[plain,noend]{algorithm2e}

\makeatletter
\renewcommand{\algocf@captiontext}[2]{#1\algocf@typo. \AlCapFnt{}#2} 
\def\@algocf@capt@plain{top}
\renewcommand{\algocf@makecaption}[2]{%
  \addtolength{\hsize}{\algomargin}%
  \sbox\@tempboxa{\algocf@captiontext{#1}{#2}}%
  \ifdim\wd\@tempboxa >\hsize
    \hskip .5\algomargin%
    \parbox[t]{\hsize}{\algocf@captiontext{#1}{#2}}
  \else%
    \global\@minipagefalse%
    \hbox to\hsize{\box\@tempboxa}
  \fi%
  \addtolength{\hsize}{-\algomargin}%
}
\makeatother

\newtheorem{theorem}{Theorem}[section]
\newtheorem{lemma}[theorem]{Lemma}

\newtheorem{Assumption}{Assumption}
\newtheorem{proposition}{Proposition}
\newtheorem{definition}[theorem]{Definition} 
\def\mcH{\mathcal{H}}
\def\E{\mathbb{E}}
\def\mcE{\mathcal{E}}
\def\P{\mathbb{P}}
\def\cH{\mathcal{H}}

\begin{document}




\title{Deep Neural Network Classifier for Multi-dimensional  Functional Data}

\author[1]{Shuoyang Wang}
\author[1]{Guanqun Cao}
\author[2]{Zuofeng Shang}

\affil[1]{Department Mathematics and Statistics, Auburn University, U.S.A.}
\affil[2]{Department of Mathematical Sciences, New Jersey Institute of Technology, U.S.A.}

\date{}

\maketitle

\begin{quotation}
\noindent \textit{Abstract:} We propose a new approach, called as functional deep neural network (FDNN), for classifying multi-dimensional functional data.
Specifically, a deep neural network is trained based on the principle components of the training data
which shall be used to predict the class label of a future data function.
Unlike the popular functional discriminant analysis approaches which rely on Gaussian assumption,
the proposed FDNN approach applies to general non-Gaussian multi-dimensional functional data.
Moreover, when the log density ratio possesses a locally connected functional modular structure,
we show that FDNN achieves minimax optimality.
The superiority of our approach is demonstrated through both simulated and real-world datasets.

\vspace{9pt} \noindent \textit{Key words and phrases:} {Functional classification}; 
 {Functional data analysis}; 
 {Functional neural networks}; 
{Minimax excess misclassification risk};
 {Multi-dimensional functional data}.
\end{quotation}

%

\pagestyle{myheadings}
\thispagestyle{plain}
{}


\section{Introduction}\label{SEC:Introduction}

Due to modern advanced technology, complex functional data are ubiquitous. A fundamental problem in functional data analysis is to 
classify a data function based on training samples.
A typical 1D example is the speech recognition data extracted from the TIMIT database, in which the training samples are digitized speech curves of American English speakers from different phoneme groups, and the task is to predict the
phoneme of a new speech curve. 
Typical 2D and 3D examples include the brain imaging data extracted from Early Mild Cognitive Impairment (EMCI)  or Alzeheimer's Disease (AD), in which
the training samples are digitized brain images, and the task is to predict the stage of a new patient.
Besides above examples, functional data classification has wide applications in various fields such as  machine learning, genetics, agriculture, chemimetrics and artificial intelligence  \citep{Song:etal:08,Leng:Muller:06,Rossi:05,Chamroukhi:12}. Recent monographs \citep{HsingEubank2015,Kokoszka:Reimherr:17}  provide comprehensive and general discussions on this field.   
   
Classical multivariate analysis techniques such as logistic regression or discriminant analysis no longer work for functional data due to its intrinsically infinite dimensionality \citep{Wang:etal:16}. 
A mainstream technique in functional data classification is based on functional principle component analysis (FPCA) such as functional discriminant analysis  \citep{Shin:08,Delaigle:etal:12,Delaigle:Hall:12, Delaigle:Hall:13,Galeano:etal:15,Dai:etal:17,Berrendero:etal:18,Park:etal:20,adrover2004globally,optfunctional2021arxiv}.
Functional discriminant analysis requires data function being Gaussian process, under which
the decision boundary is characterized by a linear or quadratic polynomial so that 
classic discriminant analysis approach can accurately recover the decision boundary.
Gaussian assumption is restrictive and often violated in practice.
When data distributions are general non-Gaussian, the resulting decision boundary
is often complicated which cannot be accurately recovered by existing approaches.
Our aim is to construct a new functional classifier to overcome this challenge.


 In this paper, we propose a new approach, called as functional deep neural network (FDNN), for multi-dimensional  functional data classification.   We start from FPCA
 to extract the functional principle components of the data functions, and then train a DNN based on these FPCs as well as their corresponding 
 class labels. As demonstrated through numerical studies, our FDNN approach performs well
 in classifying complex curve or imaging data.
Moreover, our FDNN has desirable theoretical properties.
 Intuitively, when the network architectures are suitably selected, DNN shall have large expressive power (see \cite{Petersen:Voigtlaender:2018, Yarotsky:17}) so that functional Bayes classifier can be accurately recovered, even though data distributions are complex. Specifically, we show that, when the log-ratio of the population densities demonstrates a locally connected functional modular structure, our FDNN is proven minimax optimal. 
The proposed functional modular structure is useful to overcome the infinite dimensionality of functional data,
and is meaningful as demonstrated in various examples (see Section \ref{SEC:Examples}).
Relevant modular structures have been recently adopted by researchers in nonparametric regression and classification to characterize the local behavior of the multivariate input variables, based on which DNN approaches are proven to overcome
the ``curse of dimensionality.'' 
See \cite{Schmidt:19, Bauer:Kohler:19, Liu:etal:2021, shang2021jmaa, wang2021stat, shang2021arxiv, hu2020arxiv, Kim:NN:2021, bos2021arxiv}.

 
The rest of this article is organized as follows. In Section \ref{SEC:Setup} we review functional Bayes classifier
in general setting. In Section  \ref{sec:FDNN}, we propose FDNN classifier. 
In Section \ref{SEC:Theory}, we establish
theoretical properties of FDNN under suitable technical assumptions. Section \ref{SEC:Examples} provides three progressive examples to demonstrate the validity of these technical assumptions. In Section \ref{SEC:simulation}, performances of 
FDNN and its competitors are demonstrated through simulation studies. 
In Section \ref{SEC:realdata}, we apply FDNN to speech recognition data and Alzeheimer’s Disease data. Section \ref{SEC:discussion}
summarizes the conclusions. Technical proofs are provided in Appendix and a supplement document. R programs for implementing our method are provided on
GitHub; see Section \ref{SEC:simulation}.

\section{Functional Bayes classifier under non-Gaussianity}\label{SEC:Setup}
In this section, we review functional Bayes classifier for binary classification.
Let $X(s), s\in \mathcal{S}:=[0,1]^d$ be a random process with $\int_{\mathcal{S}}\mathbb{E}X(s)^2 ds<\infty$,
and $Y\in\{-1,1\}$ be a uniform random class label such that,
under $Y=k$, $X(s)$ has unknown mean function $\mu_k(s)$ and unknown covariance function $\Omega_k(s,s')$, for $s,s'\in\mathcal{S}$. 
Suppose that $\Omega_k$ satisfies a Karhunen–Lo\'{e}ve decomposition:
\begin{equation}\label{eqn:Omegak}
\Omega_k(s,s')=\sum_{j=1}^\infty\lambda_{kj}\psi_{kj}(s)\psi_{kj}(s'), s,s'\in\mathcal{S},
\end{equation}
where $\psi_{kj}, j\ge1$ is an orthonormal basis of $L^2(\mathcal{S})$ with respect to the usual $L^2$ inner product,
and $\lambda_{k1}\ge \lambda_{k2}\ge \cdots>0$ are nonincreasing positive eigenvalues.  
Notably, (\ref{eqn:Omegak}) requires the covariance functions being decomposed in terms of the same eigenfunctions, 
which is a common assumption in functional classification literature; see \citep{Delaigle:Hall:12} and \cite{Dai:etal:17}.
Further relaxation of this assumption is discussed in Section  .

Under $Y=k$, write $X(s)=\sum_{j=1}^\infty\xi_j\psi_{kj}(s)$, where
$\xi_j$'s are  pairwise uncorrelated random coefficients.
Let $\bs{\xi} =(\xi_1 ,\xi_2 ,\ldots)$ and 
$h_{k}(\cdot)$ be the unknown conditional density of $\bs{\xi}$ under $Y=k$.
Define $Q^\ast(\cdot)$ as the log density ratio functional between the two classes: 
$$
Q^\ast(\bs{\xi})=\log \left( \frac{h_{1}(\bs\xi)}{h_{-1}(\bs\xi)}\right).
$$
The functional Bayes rule
for classifying a data function $X \in L^2(\mathcal{S})$ thus has an expression
\begin{eqnarray}\label{DEF:FBAYES}
G^\ast(X)=
\begin{cases}
      1, & Q^\ast(\bs{\xi})\geq 0,\\
      -1, & Q^\ast(\bs{\xi})< 0.
   \end{cases}
\end{eqnarray}
Direct estimation of $Q^\ast$ is infeasible due to the infinite dimensionality of the input. A common practice is to estimate its finite-dimensional truncation.
For $J\ge1$, let $\bs{\xi}_J=(\xi_1,\ldots,\xi_J)^\top$ be the leading $J$ components of $\bs{\xi}$ and $h_k^{(J)}(\cdot)$ be the marginal density of $\bs{\xi}_J$ under $Y=k$, for $k=\pm1$. Define the truncated log density ratio 
\[
Q^\ast_J(\bs{\xi}_J) =  \log\left(\frac{h_1^{(J)}(\bs{\xi}_J)}{h_{-1}^{(J)}(\bs{\xi}_J)}\right),
\]
which is the log density ratio of $h_1^{(J)}$ to $h_{-1}^{(J)}$. The intuition is that, when $J$ is large, $h_k^{(J)}$ approaches $h_k$ so that $Q^\ast_J$ is an accurate approximation of $Q^\ast$.
Our aim is to design an efficient method to estimate $Q^\ast_J$,
which will in turn estimate $Q^\ast$.
\section{Functional deep neural network classifier}\label{sec:FDNN}
Suppose we observe $n$ i.i.d. training samples $\{(X_i (s), Y_i): 1\le i\le n, s\in\mathcal{S}\}$,
which are independent of $X(s),s\in\mathcal{S}$ to be classified.
For $k=\pm1$, define sample covariance function
\[
\widehat{\Omega}_k(s,s')=\frac{1}{n_k}\sum_{i\in I_k}(X_i(s)-\bar{X}_k(s))(X_i(s')-\bar{X}_k(s')),\,\,\,\,
s,s'\in\mathcal{S},
\]
where $I_k$ is the collection of $i$ such that $Y_i=k$,
$n_k:=|I_k|$ and $\bar{X}_k(s)=\frac{1}{n_k}\sum_{i\in I_k}X_i(s)$
is the sample mean function of class $k$.
Perform Karhunen–Lo\'{e}ve decomposition for $\widehat{\Omega}_k$:
\[
\widehat{\Omega}_k(s,s')=\sum_{j=1}^\infty\widehat{\lambda}_{kj}\widehat{\psi}_{kj}(s)\widehat{\psi}_{kj}(s'), s,s'\in\mathcal{S},
\]
and write the sample data function $X_i$, under $Y_i=k$, as
\[
X_i(s)=\sum_{j=1}^\infty\widehat{\xi}_{ij}\widehat{\psi}_{kj}(s), i=1,\ldots,n.
\]
Intuitively, $\widehat{\bs\xi}^{(i)}:=(\widehat{\xi}_{i1},\widehat{\xi}_{i2},\ldots)$ is an estimator of $\bs{\xi}^{(i)}:=(\xi_{i1},\xi_{i2},\ldots)$, in which
$\xi_{ij}$ are unobservable random coefficients of $X_i$ with respect to the population basis $\psi_{kj}$. Hence, it is natural to design classifiers based on $\widehat{\bs\xi}^{(i)}$'s. 

Let $\widehat{\bs\xi}^{(i)}_{J}=(\widehat{\xi}_{i1},\ldots,\widehat{\xi}_{iJ})^\top$ be the $J$-dimensional truncation of $\widehat{\bs\xi}^{(i)}$ for $i=1,\ldots,n$. When $X_i$'s are Gaussian processes, various classifiers have been proposed such as centroid classifier \citep{Delaigle:Hall:12}, QDA \citep{Delaigle:Hall:13} and nonparametric Bayes classifier \citep{Dai:etal:17}. 
When $X_i$'s are non-Gaussian,
one major challenge is the underlying complicated form of the conditional densities $h_1$ and $h_{-1}$ so that estimation of $Q_J^\ast$ is typically difficult.
Inspired by the rich approximation power of DNN, in this section, we propose a new classifier called FDNN (functional+DNN) that can accurately estimate Bayes classifiers even when $h_1$ and $h_{-1}$ are non-Gaussian complicated.

 We will train a DNN to estimate $Q^\ast_J$ based on $\widehat{\bs\xi}^{(i)}_{J}$'s. 
In what follows, we will describe our method in details. 
Let $\sigma$ denote the rectifier linear unit (ReLU) activation function,  i.e.,  $\sigma(x)=(x)_+$ for $x\in\mathbb{R}$.
For any real vectors $\bs{V}=(v_1,\ldots,v_w)^\top$ and $\bs{y}=(y_1,\ldots,y_w)^\top$, define the shift activation function $\sigma_{\bs{V}}(\bs{y})=(\sigma(y_1-v_1),\ldots,\sigma(y_w-v_w))^\top$.  
For $L\ge1$, $\bs{p}=(p_1,\ldots,p_L)\in\mathbb{N}^L$, 
let $\mathcal{F}(L,J,\bs{p})$ denote the class of fully connected feedforward DNN with $J$ inputs, $L$ hidden layers and, for $l=1,\ldots,L$, $p_l$ nodes on the $l$th hidden layer.
Equivalently, any $f\in\mathcal{F}(L,J,\bs{p})$ has an expression
 \begin{equation} \label{EQ:f}
f(\bs x) = \mathbf{W}_L\sigma_{\bs{V}_L} \mathbf{W}_{L-1}\sigma_{\bs{V}_{L-1}}\ldots \mathbf{W}_1\sigma_{\bs{V}_1} \mathbf{W}_0\bs x, \,\,\,\, {\bs x}\in\mathbb{R}^J,
\end{equation}
where $\mathbf{W}_l\in\mathbb{R}^{p_{l+1}\times p_{l}}$, for $l=0,\ldots,L$, are weight matrices, $\bs{V}_l\in\mathbb{R}^{p_l}$, for $l=1,\ldots,L$,
are shift vectors. Here we adopt the convention that $p_0=J$ and $p_{L+1}=1$.


We consider the following class of DNN:
\begin{eqnarray}\label{EQ:class}
\hspace{3mm} \mathcal{F}(L, J, \bs{p},   B)
 = \left\{ f\in \mathcal{F}(L, J, \bm{p}) :  \max_{0\le l\le L}\| \mathbf{W}_l\|_{\infty}\le B, \max_{1\le l\le L}\|\mathbf{v}_l\|_{\infty} \leq B\right\}, \nonumber
\end{eqnarray}
where $ \| \cdot\|_{\infty}$ denotes the maximum-entry norm of a matrix/vector
or supnorm of a function,  and  $B>0$ controls the largest weights and shifts.
 


Given the training data
$(\bs\xi^{(1)}_J, Y_1),\ldots,(\bs\xi^{(n)}_J, Y_n)$, let
\begin{equation}\label{hatf:FDNN}
\widehat{f}_{\phi}(\cdot) = \arg\min_{f\in \mathcal{F}(L, J, \bs{p}, B)} \frac{1}{n}\sum_{i=1}^{n }\phi(f(\widehat{\bs\xi}^{(i)}_{J})Y_i),
\end{equation}
where $\phi(x) = \max(1-x, 0)$ denotes the hinge loss.
We then propose the following FDNN classifier:
for $X\in L^2(\mathcal{S})$,
\begin{eqnarray}\label{DEF:classifier:dnn}
 {\widehat{G}^{FDNN}(X)}=
\left\{\begin{array}{cc}
1, & \widehat{f}_{\phi}(\bs\xi_J)\ge 0,\\
-1, & \widehat{f}_{\phi}(\bs\xi_J)< 0.
\end{array}\right.
\end{eqnarray}

In practice, we suggest the following data-splitting method for selecting
$(L, J, \bs{p}, B)$:
\begin{itemize}
    \item Step 1. Randomly divide the whole sample $(\widehat{\bs\xi}^{(i)}_{J},Y_i)$'s into two subsets indexed by $\mathcal{I}_1$ and $\mathcal{I}_2$, respectively, with about $|\mathcal{I}_1|=0.8n$ and $|\mathcal{I}_2|=0.2n$.
    \item Step 2. For each $(L,J,\bs{p},B)$, we train a DNN $\widehat{f}_{L,J,\bs{p},B}$ using (\ref{hatf:FDNN}) based on subset $\mathcal{I}_1$, and then calculate the testing error based on subset $\mathcal{I}_2$ as\vspace{-3mm}
\begin{equation}\label{p3:testing:error}
\textrm{err}(L,J,\bs{p},B)=\frac{1}{|\mathcal{I}_2|}\sum_{i\in\mathcal{I}_2}I(\widehat{f}_{L,J,\bs{p},B}(\widehat{\bs\xi}^{(i)}_{J})Y_i<0).\vspace{-3mm}
\end{equation}
\item Step 3. Choose $(L,J,\bs{p},B)$, possibly from a preselected set, to minimize $\textrm{err}(L,J,\bs{p},B)$.
\end{itemize}



\section{Minimax optimality of FDNN}\label{SEC:Theory}

For a generic functional classifier $\widehat{G}$, 
its excess misclassification risk is defined as
$\mcE_h(\widehat{G}):=\E[R_h(\widehat{G})-R_h(G^\ast)]$,
where $R_h(\widehat{G}):=\mathbb{E}_{h}[\mathbb{I}\{\widehat{G}(X)\neq Y\}]$ is the misclassification risk of $\widehat{G}$ taken with respect to $(X,Y)$ under $h:=\{h_1,h_{-1}\}$,
with $Y$ the true class label of $X$.
A central task is to design $\widehat{G}$ that achieves minimax excess misclassification risk (MEMR), i.e.,
\begin{equation}\label{eqn:memr}
\max_{h\in\mcH}\mcE_h(\widehat{G})\asymp \inf_{\widehat{G}}\max_{h\in\mcH}\mcE_h(\widehat{G}),
\end{equation}
where $\mcH$ is a proper class of $h$ to be described later and the infimum is taken over all classifiers based on training samples. 
Classifiers satisfying (\ref{eqn:memr}) are called as minimax optimal.

There is a rich literature on construction of minimax optimal classifiers when data dimension is fixed or diverging.
For instance, classic nonparametric approaches, such as ones directly estimating Bayes classifier nonparametrically, are proven minimax optimal in fixed-dimension regime 
  \citep {Mammen:etal:99, Tsybakov:04, Tsybakov:09, Lecue:08, Galeano:etal:15, Farnia:Tse:16, Mazuelas:etal:20, hu2020arxiv}.
When data are high-dimensional Gaussian, discriminant analysis approaches are proven minimax optimal  \citep{Cai:Zhang:19, Cai:Zhang:19b}.
On the other hand, under functional Gaussian data, researchers have proposed various functional classifiers, including functional quadratic discriminant analysis (FQDA)  \citep{Shin:08,Delaigle:etal:12,Delaigle:Hall:12, Delaigle:Hall:13,Galeano:etal:15,Dai:etal:17,Berrendero:etal:18,Park:etal:20,Cai:Zhang:19, Cai:Zhang:19b}. Gaussianity leads to a linear or quadratic polynomial $Q^\ast$
which can be effectively estimated by FQDA, based on which 
\cite{optfunctional2021arxiv} showed that FQDA is minimax optimal.
It is still unclear how to design optimal functional classifiers when data are non-Gaussian,
a gap that the present article attempts to close.

In this section, we will establish minimax optimality of FDNN classifier under non-Gaussian functional data. {For technical convenience, 
assume that the two populations have common known basis, i.e., $\psi_{+1j}(\cdot)=\psi_{-1j}(\cdot)$.} Therefore, we can train FDNN classifier based on $\bs{\xi}^{(i)}_J:=(\xi_{i1},\ldots,\xi_{iJ})^\top$, for $i=1,\ldots,n$. 
We will first derive an upper bound for the excess misclassification risk of our FDNN classifier,
and then derive a lower bound for the MEMR which matches the above upper bound. Therefore, our FDNN is able to achieve sharp rate of MEMR.
Extensions to general basis are possible with more tedious technical arguments.

Before proceeding further, we introduce some technical assumptions. 
At high levels, our assumptions are different from those proposed under Gaussian case.
For instance, in either high- or infinite-dimensional Gaussian data,
it is well known that density ratio between two Gaussian population densities
has an explicit expression in terms of mean difference and variance ratio, which impacts the sharp rate of MEMR. 
More precisely, in high-dimensional Gaussian data classification, the rate depends on the number of nonzero components of mean difference vector \citep{Cai:Zhang:19, Cai:Zhang:19b};
in Gaussian functional data classification, the rate depends on the decay orders
of both mean difference series and variance ratio series \citep{optfunctional2021arxiv}.
Nonetheless, in general non-Gaussian case, likelihood ratio does not have an explicit expression, therefore, one cannot simply use mean or variance discrepancy to characterize the sharp rate of MEMR. 

In traditional non-Gaussian multivariate data classification, a common strategy is to assume smooth density ratio and controllable noise, under which minimax optimal classifiers were proposed (see \cite{Mammen:etal:99,Tsybakov:04,Audibert:Tsybakov:07,Kim:NN:2021} and references therein).
In the functional data framework, the input variable of $Q^\ast$ is infinite-dimensional, hence, the above strategy no longer works.
We instead propose a set of functional conditions on $Q^\ast$ under which minimax optimality shall be established.
Such conditions are viewed as infinite-dimensional extensions of \cite{Audibert:Tsybakov:07} and \cite{Schmidt:19}.

For $t\ge1$, a measurable subset $D\subset \mathbb{R}^t$ and constants $\beta,K>0$,
define 
\begin{eqnarray*}
&&\mathcal{C}^\beta(D, K)\\
&=&\left\{f:D\mapsto\mathbb{R}\big|\sum_{\bs{\alpha}:|\bs{\alpha}|<\beta} \| \partial^{\bs{\alpha}}f\|_{\infty} + \sum_{\bs{\alpha}:|\bs{\alpha}|=\lfloor{\beta}\rfloor}\sup_{\bs{x}, \bs{x}' \in D, \bs{x} \neq \bs{x}' } \frac{|\partial^{\bs{\alpha}}f(\bs{x}) - \partial^{\mathbf{\alpha}}f(\bs{x}') |}{\|\bs{x} - \bs{x}'\|_{\infty}^{\beta - \lfloor\beta\rfloor}} \leq K\right\},
\end{eqnarray*}
where $\partial^{\bs{\alpha}}$ = $\partial^{\alpha_1}\ldots\partial^{\alpha_t}$ denotes the
partial differential operator with multi-index $\bs{\alpha}$ = $(\alpha_1, \ldots, \alpha_t) \in \mathbb{N}^t$, $|\bs{\alpha}|=\alpha_1+\cdots+\alpha_t$.
Equivalently, $\mathcal{C}^\beta(D, K)$ is the ball of $\beta$-H\"{o}lder smooth functions 
on $D$ with radius $K$.
A function $f: \mathbb{R}^t\to\mathbb{R}$ is said to be locally 
$\beta$-H\"{o}lder smooth if 
for any $a,b\in\mathbb{R}$, there exists a constant $K$ (possibly depending on $a,b$) such that $f\in\mathcal{C}^\beta([a,b]^t,K)$.

For $q\ge0, J\ge 1$, let $d_0=J$ and $d_{q+1}=1$. For $\bs{d}=(d_1, \ldots, d_q)\in\mathbb{N}_+^{q}$,  $\bs{t}= (t_0, \ldots, t_q)\in\mathbb{N}_+^{q+1}$ with $t_u\le d_u$ for $u=0,\ldots, q$, 
$\bs{\beta} := (\beta_0, \ldots, \beta_q)\in\mathbb{R}_+^{q+1}$,
let $\mathcal{G}(q, J, \bs d, \bs t, \bs\beta)$ be the class of functions $g$ 
satisfying a modular expression
\begin{equation}\label{modular:expression}
g(\bs{x})=g_q \circ \cdots \circ g_0(\bs{x}),\,\,\forall \bs{x}\in\mathbb{R}^{d_0},
\end{equation}
where $g_u=(g_{u1},\ldots,g_{ud_{u+1}}): \mathbb{R}^{d_u}\mapsto\mathbb{R}^{d_{u+1}}$
and $g_{uv}: \mathbb{R}^{t_u}\mapsto\mathbb{R}$ are locally $\beta_u$-H\"{o}lder smooth.
The $d_u$ arguments of $g_u$ are locally connected
in the sense that each component $g_{uv}$ only relies on $t_u (\le d_u)$ arguments.
Similar structures have been considered by \cite{Schmidt:19, Bauer:Kohler:19, Liu:etal:2021, shang2021jmaa, wang2021stat, shang2021arxiv, hu2020arxiv, Kim:NN:2021, hu2020arxiv}
in multivariate regression or classification to overcome high-dimensionality. Generalized additive model \citep{gam1990} and tensor product space ANOVA model \citep{lin2000aos}
are special cases; see \cite{Liu:etal:2021}.

Let $\mathcal{H}^\ast\equiv\mathcal{H}^\ast\left(q, \bs{d}, \bs{t}, \bs{\beta}\right)$ be the class of population densities $h=\{h_1,h_{-1}\}$ of $\bs{\xi}$ such that, for any $J\ge1$, $Q_J^\ast\in\mathcal{G}\left(q,  J, \bs{d}, \bs{t}, \bs{\beta}\right)$.
Equivalently, for any $h\in \mathcal{H}^\ast$ and $J\ge1$, the corresponding truncated log density ratio $Q_J^\ast$ has a modular structure (\ref{modular:expression}) with certain smoothness. Although $Q_J^\ast$ has $J$ arguments, it involves at most $t_0d_1$ effective arguments, implying that the two population densities differ by a small number of variables. Relevant conditions are necessary for high-dimensional classification. For instance, 
in high-dimensional Gaussian data classification, \cite{Cai:Zhang:19, Cai:Zhang:19b} show that, to consistently estimate Bayes classifier, it is necessary 
that the mean vectors differ at a small number of components.
The modular structure holds for arbitrary $J$, which may be viewed as a functional extension of \cite{Schmidt:19}.
Note that the density class $\mathcal{H}^\ast$ covers many popular models
studied in literature, either Gaussian or non-Gaussian; see Section \ref{SEC:Examples}.
Moreover, we introduce the following regularity conditions on $Q^\ast$.

\begin{Assumption}(Functional Tsybakov noise condition)\label{A3} 
There exist constants $C>0$ and $\alpha\geq 0$ such that 
\begin{equation}
    \P\left(\left|\frac{1-\exp\left\{ -Q^\ast\left( \bs\xi\right)\right\}}{1+\exp\left\{ -Q^\ast\left( \bs\xi\right)\right\}} \right| \le x\right)\le Cx^{\alpha}, \;\;\;\;\;\;\forall x>0.
\end{equation}
\end{Assumption}
\begin{Assumption}(Approximation error of $Q^\ast_J$)\label{A4}
There exist a constant $J_0\ge1$ and decreasing functions $\epsilon(\cdot):\left[1, \infty\right)\to\mathbb{R}_+$ and $\Gamma(\cdot):\left[0, \infty\right)\to\mathbb{R}_+$,
with $\sup_{J\ge1}J^\varrho\epsilon(J)<\infty$ for some $\varrho >0$ and $\int_0^\infty\Gamma(x)dx<\infty$, such that for any $J\ge J_0$ and $x>0$,
\begin{equation}
    \P\left(|Q^\ast(\bs\xi)-Q^\ast_J(\bs\xi_J) | \geq x\right) \leq \epsilon(J)\Gamma(x).
\end{equation}
\end{Assumption}
Assumption \ref{A3} characterizes the discrepancy between $Q^\ast$ and random guess.
Specifically, it requires that the probability of $Q^\ast$ close to $0$
by $x$ is upper bounded by an order $x^\alpha$. 
Assumption \ref{A3} is a functional extension of the classic \textit{Tsybakov noise condition}, which is necessary in establishing minimax classification in multivariate case  \citep{Mammen:etal:99,Tsybakov:04}).
Assumption \ref{A4} provides an upper bound on the probability of $Q^\ast$ differing from $Q^\ast_J$ by at least $x$, which approaches zero if either $J$ or $x$ tends to infinity,
implying that $Q^\ast_J$ is an accurate approximation of $Q^\ast$. Both assumptions can be verified in several concrete examples included in   Section \ref{SEC:Examples}.

Our MEMR results will be based on the following class of population densities of $\bs{\xi}$:
\begin{eqnarray*}
\cH &\equiv& \cH\left(q, \bs{d}, \bs{t}, \bs{\beta}, \alpha, C, \epsilon(\cdot), \Gamma(\cdot)\right)
= \left\{h\in\cH^\ast: \text{$Q^\ast$ satisfies both Assumptions  \ref{A3} and  \ref{A4}}\right\}.\nonumber
\end{eqnarray*}

Finally, we introduce an assumption on the orders of $(L,J,\bs{p},B)$, under which the exact rate of MEMR shall be established. Let 
\[
S_0=\min_{0\le u\le q}\frac{\beta_u^\ast(\alpha+1)}{\beta_u^\ast(\alpha+2) + t_u}, S_1=\max_{0\le u\le q} \frac{t_u}{\beta_u^\ast(\alpha+2) + t_u}, S_2=\min_{0\le u\le q} \frac{1}{\beta_u^\ast(\alpha+2) + t_u},
\] 
where $\beta_u^{\ast} := \beta_u \prod_{k=u+1}^q (\beta_k \wedge 1)$.

\begin{Assumption}\label{A5}
The DNN class $\mathcal{F}(L,J,\bs{p},B)$ satisfies 
\begin{enumerate}[label=(\alph*)]
    \item\label{A5:a} $L\asymp \log n$; 
    \item\label{A5:b} $\left( n\log^{-3} n\right)^{S_0/\rho} \lesssim J \lesssim (n\log^{-3} n)^{S_1}$; 
    \item\label{A5:c} $ \max_{1\leq \ell \leq L} p_\ell\asymp (n\log^{-3} n)^{S_1}$;
    \item\label{A5:d} $B\asymp (n\log^{-3} n)^{S_2}$.
\end{enumerate}
\end{Assumption}
Assumption \ref{A5}\ref{A5:a}, \ref{A5:c}, and \ref{A5:d} provide exact orders on $L,\bs{p}, B$, respectively.
Assumption \ref{A5}\ref{A5:b} provides a range on $J$. Notably,
this condition implies $\varrho\ge S_0/S_1$, i.e., the function $\epsilon(J)$ rapidly converges to zero when $J\to\infty$.


 
\begin{theorem}\label{THM: minimax}
There exist positive constants $C_1, C_2$, depending on $q, \bs{d}, \bs{t}, \bs{\beta}, \alpha, C, \epsilon(\cdot)$, and $\Gamma(\cdot)$, such that the following results hold:
\begin{enumerate}[label=(\roman*)]
\item\label{Thm:lower} $\inf_{\widehat{G}}\sup_{h\in\cH}\mathcal{E}_h(\widehat{G})\ge C_1 n^{-S_0}$,
where the infimum is taken over all classifiers $\widehat{G}$ based on training samples;
\item\label{Thm:upper} under Assumption \ref{A5}, it holds that
$$\sup_{h \in \cH} \mathcal{E}_h(\widehat{G}^{FDNN})\le C_2\left(\frac{\log^3 n}{n}\right)^{S_0}.$$
\end{enumerate}
\end{theorem}
Theorem \ref{THM: minimax} establishes a nonasymptotic rate for the MEMR which is of order $n^{-S_0}$. Moreover, the proposed FDNN classifier is able to achieve this rate up to a logarithmic factor, and hence, is minimax optimal. Since $S_0$ involves the intrinsic dimensions $t_u$'s rather than the original dimensions $d_u$'s, the rate of MEMR is typically fast, demonstrating the theoretical advantage of our FDNN classifier. 
 
\section{Examples}\label{SEC:Examples}
The minimax results in Section \ref{SEC:Theory} are based on parameter space $\cH$.
In this section, we provide some concrete examples to demonstrate the validity of such space. 

\subsection{Gaussian functional data with independent coefficients}\label{sec:gaussian:fd}
Suppose that, under $Y=k$, the random coefficients $\xi_j$ are independent Gaussian with mean $\mu_{kj}$ and variance $\lambda_{kj}$. Define $M=\left\{j: \mu_{1j}\neq \mu_{-1j}\right\}$ and $N=\{j: \lambda_{1j}\neq\lambda_{-1j}\}$. Assume that $M,N$ are mutually disjoint with common cardinality $\omega$. It can be shown that, for any $J\ge J_0:=\max M\cup N$,
$Q^\ast_J(\bm{\xi}_J)=g_1(g_0(\bs \xi_J))$,
where $g_0$ has components $g_{0j}(\xi_j) = a_j \xi_j^2 + b_j \xi_j + c_j$ for some constants $a_j, b_j, c_j$ depending on $\mu_{1j}$, $\mu_{-1j}$, $\lambda_{1j}$, $\lambda_{-1j}$, and
$g_1(g_0(\bs \xi_J))=\sum_{j\in M\cup N} g_{0j}(\xi_j)$.
Clearly, $d_0=J$ and $t_0=1$, and $M\cup N$ has cardinality $2\omega$,
$d_1=t_1=2\omega$. So $Q^\ast_J\in\mathcal{G}(1,J,2\omega,(1,2\omega),\beta)$ for any $\beta>0$.
Meanwhile,  since $Q_J^\ast = 0$ for all $J\geq J_0$, and for any function $\epsilon(\cdot)$ with exponential tails and any density $\Gamma(\cdot)$, Assumption \ref{A3} holds for $\alpha=1$  and Assumption \ref{A4} holds for $J_0$. 

\subsection{Student's $t$ functional data with independent coefficients}\label{sec:student:t:fd}
Suppose that, under $Y=k$, $\xi_j$ are independent Student's $t$ variables $t_{\nu_{kj}}$, where $\nu_{kj}\ge 1$ are degrees of freedom of the $t$ variables. Define $M=\left\{j: \mu_{1j}\neq\mu_{-1j}\right\}$ whose cardinality is $\omega$. It can be shown that, for any $J\ge J_0:=\max M\cup N$,
$Q^\ast_J(\bm{\xi}_J)=g_1(g_0(\bm{\xi}_J))$, where $g_0$ has components 
\[
g_{0j}(\xi_j) = \log e_j - \frac{\nu_{1j}+1}{2}\log\left(1+\frac{\xi_j^2}{\nu_{1j}}\right) + \frac{\nu_{-1j}+1}{2}\log\left(1+\frac{\xi_j^2}{\nu_{-1j}}\right),
\]
for some constant $e_j$ depending on $\nu_{kj}$, and $g_1(g_0(\bs \xi_J))=\sum_{j\in M\cup N} g_{0j}(\xi_j)$. Similar to Section \ref{sec:gaussian:fd}, we have 
$Q_J^\ast\in \mathcal{G}(1,J,2\omega,(1,2\omega),\beta)$ for any $\beta>0$.
Assumptions \ref{A3} and \ref{A4} can be similarly verified as well.




\subsection{Student's $t$ functional data with dependent coefficients}\label{sec:multi:student:t:fd}
We consider an extension of Section \ref{sec:student:t:fd} which involves dependent coefficients. Let $p\ge1$ and $\nu \geq 2$ be integers.
Suppose that, under $Y=k$, $\bs\zeta_j:= \left( \xi_{j}, \xi_{j+1}, \ldots, \xi_{j+p-1} \right)^\top$, $j=1,p+1,2p+1,\ldots$ are independent multivariate Student's $t$ vectors following $t_\nu(\bs\mu_{kj}, \bs\Sigma_{kj})$, where $\bs\mu_{kj}\in\mathbb{R}^p$ and positive definite $\bs\Sigma_{kj}$ is $p\times p$ positive definite.
Define $M=\left\{j: \bs\mu_{1j}\neq \bs\mu_{-1j}\right\}$ and $N=\left\{j: \bs\Sigma_{1j}\neq\bs\Sigma_{-1j}\right\}$. Assume sets $M$ and $N$ are mutually disjoint with common cardinality $\omega$.
For any $J\ge J_0:=J_0:=\max M\cup N+p-1$, then it can be shown that
\begin{eqnarray*}
&&Q^\ast_J(\bm{\xi}_J)\\
&=& \sum_{j\in M \cup  N} \left\{\frac{1}{2}\log\left(\frac{|\bs\Sigma_{-1j}|}{|\bs\Sigma_{1j}|}\right)^{1/2}
+\frac{\nu+p}{2}\log\left(\frac{1+\nu^{-1}(\bs\zeta_j - \bs\mu_{-1j})^{\top}\bs\Sigma_{-1j}^{-1}(\bs\zeta_j - \bs\mu_{-1j})}{1+\nu^{-1}(\bs\zeta_j - \bs\mu_{1j})^{\top}\bs\Sigma_{1j}^{-1}(\bs\zeta_j - \bs\mu_{1j})}\right)\right\}.
\end{eqnarray*}
Note that there are $2\omega$ terms in the above sum. Similar to Sectoions \ref{sec:gaussian:fd}, we have $Q_J^\ast\in \mathcal{G}(1,J,2\omega,(1,2\omega),\beta)$ for any $\beta>0$. Assumptions \ref{A3} and \ref{A4} can be similarly verified as well.

\section{Simulation study}\label{SEC:simulation}
In this section, we examine the performances of FDNN and two competitors,
quadratic discriminant method (QD)  proposed in \cite{Delaigle:Hall:13} and the nonparametric Bayes classifier (NB) proposed in \cite{Dai:etal:17}, through simulation studies. Our studies involve both $d=1$ and $d=2$, corresponding to 1D and 2D functional data, respectively. All experiments are conducted in \texttt{R}.   We summarize R codes and examples for the proposed FDNN algorithms   on \texttt{GitHub}  (\url{https://github.com/FDASTATAUBURN/fdnn-classification}).

For 1D functional data, we considered two data generation processes (DGP).
\begin{itemize}
\item\textit{DGP1:} Generate $X(s)=\sum_{j=1}^{3} \xi_{j} \psi_j(s)$, $s\in[0,1]$, where $\psi_1(s)=\log(s+2)$, $\psi_2(s)=s$ and $\psi_3(s)=s^3$. 
 Under class $k$, generate independently $(\xi_1,\xi_2,\xi_3)^{\top}\sim N(\bm{\mu}_k,\bm{\Sigma}_k)$, where
$\bs\mu_1= \left(-1, 2, -3 \right)^\top$, $\bs\Sigma_1= \text{diag}\left( \frac{3}{5}, \frac{2}{5}, \frac{1}{5}\right) $,   $\bs\mu_{-1}= \left(-\frac{1}{2}, \frac{5}{2}, -\frac{5}{2} \right)^\top$,  $\bs\Sigma_{-1}=\text{diag}\left( \frac{9}{10}, \frac{1}{2}, \frac{3}{10}\right) $. 

\item\textit{DGP2:} Generate $X(s)=\sum_{j=1}^{3} \xi_{j} \psi_j(s)$, $s\in[0,1]$, where $\psi_j(s)$'s are the same as in DGP1.
Under class $1$,
generate independently $(\xi_1,\xi_2,\xi_3)^{\top}\sim N(\bm{\mu}_1,\bm{\Sigma}_1)$,
where
$\bs\mu_1= \left(-1, 2, -3 \right)^\top$, $\bs\Sigma_1= \text{diag}\left(3, 2, 1\right)$; under class $-1$, generate independently $\xi_{j}\sim t_{7-2j}$, $j=1,2,3$. 
\end{itemize}

For 2D functional data, we considered two DGPs:
\begin{itemize}
\item\textit{DGP3:} Generate $X(s_1,s_2)= \sum_{j=1}^{4} \xi_{j}\psi_j(s_1,s_2)$, $0\le s_1,s_2\le1$, where $\psi_1(s_1, s_2)=s_1s_2$, $\psi_2(s_1, s_2)=s_1s_2^2$, $\psi_3(s_1, s_2)=s_1^2s_2$, $\psi_4(s_1, s_2)=s_1^2s_2^2$. Under class $k$, generate independently $(\xi_1,\xi_2,\xi_3,\xi_4)^{\top}\sim N(\bm{\mu}_k,\bm{\Sigma}_k)$,  
where $\bm\mu_1=(8,-6,4,-2)^\top$,
$\bs\Sigma_1= \text{diag}\left( 8, 6, 4, 2\right) $,   $\bs\mu_{-1}= \left(-\frac{7}{2}, -\frac{5}{2}, \frac{3}{2},  -\frac{1}{2}\right)^\top$,  $\bs\Sigma_{-1}=\text{diag}\left( \frac{9}{2}, \frac{7}{2}, \frac{5}{2}, \frac{3}{2}\right)$.

\item \textit{DGP4:} Generate $X(s_1,s_2)= \sum_{j=1}^{4} \xi_{j}\psi_j(s_1,s_2)$, $0\le s_1,s_2\le1$, where $\psi_j(s_1, s_2)$'s are the same as in {DGP3}. For $j=1,2,3,4$, under class 1, generate independently ${\xi_j}\sim t_{2j}(  0)$; under class $-1$, generate independently ${\xi_j}\sim t_{2j+1}( \mu_j)$, with non-central parameter $\mu_1=2$, $\mu_2=\frac{3}{2}$, $\mu_3=1$, and $\mu_4=\frac{1}{2}$. 
\end{itemize}
In each DGP, we generated $n$ training data functions and 500 testing data functions, with $n=40,100,200,400$. Each data function was sampled over $50$ grid points in the respective domain.
Misclassification rates were evaluated based on $100$ replicated datasets. Network parameters were selected based on training data using Steps 1-3 in Section \ref{sec:FDNN}.
Table \ref{TAB:S1} and Table \ref{TAB:sim_2d_DNN} summarizes the misclassification rates with standard deviations for 1D and 2D functional data, respectively. Specifically, the proposed FDNN outperforms QDA and NB in all settings. Though FDNN has larger standard deviation than QD and NB,
the values decrease when $n$ becomes large. As no existing alternative methods for 2D functional data classification available, we only present the performance of the proposed FDNN classifier. We observe that the misclassification risk decreases as the sample size increase for both Gaussian and non-Gaussian functional data.

\begin{table}
\begin{tabular}{cccccccc}
\hline
 \multirow{2}{*}{$n$}&& DGP1 && &&DGP2& \\
\cline{2-4}   \cline{6-8}
 &  \multicolumn{1}{c}{FDNN} & \multicolumn{1}{c}{QD} & \multicolumn{1}{c}{NB}& &\multicolumn{1}{c}{FDNN} & \multicolumn{1}{c}{QD} & \multicolumn{1}{c}{NB}\\
\cline{1-4}   \cline{6-8}
40    & 31.76(0.10)  & 38.58(0.02) & 38.33(0.02)& &16.69(0.04)  & 39.99(0.01) & 39.26(0.03)\\
100   & 18.82(0.10)  & 37.91(0.02) & 41.03(0.02)& &13.20(0.01)  & 38.42(0.09) & 40.27(0.03)\\
200   & 13.19(0.10)  & 37.35(0.02) & 39.92(0.02)& &12.29(0.01)  & 42.63(0.02) & 39.84(0.04)  \\
400   & 9.62(0.04)   & 36.75(0.02) & 38.54(0.02)& &12.40(0.01)  & 43.98(0.09) & 38.51(0.04)  \\
\hline
\end{tabular}
\caption{\label{TAB:S1}Misclassification rates ($\%$) with standard errors in brackets for DGP1 and DGP2}
\end{table}



\begin{table}
\centering
\begin{tabular}{ccccc}
\hline
  & \multicolumn{4}{c}{n}\\
\cline{2-5} 
  &  $40$ & $100$  &$200$ &$400$  \\
\hline
DGP 3  & 0.170(0.066) & 0.148(0.055) & 0.139(0.054) &  0.127(0.040) \\
DGP 4  & 0.139(0.055) & 0.127(0.014) & 0.127(0.040) &  0.123(0.011)  \\
\hline
\end{tabular}
\caption{\label{TAB:sim_2d_DNN}Misclassification rates ($\%$) with standard errors in brackets for DGP3 and DGP4 with the FDNN classifier}
\end{table}

\section{Real data illustrations}\label{SEC:realdata}
 
 \subsection{TIMIT database}\label{SEC:Real_1D}
\begin{figure}
\begin{center}
\includegraphics[width=6.5cm, height=5cm]{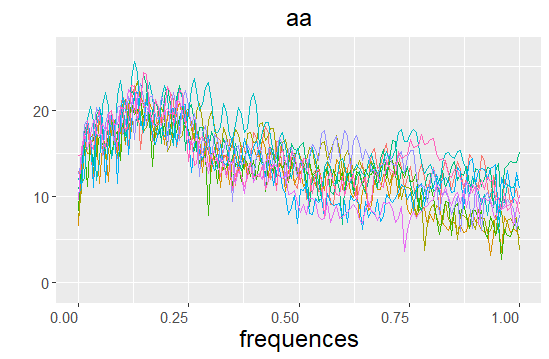}
\includegraphics[width=6.5cm, height=5cm]{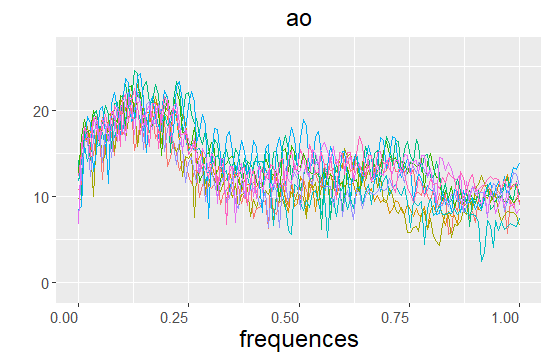}
\vskip 0.5cm
\includegraphics[width=6.5cm, height=5cm]{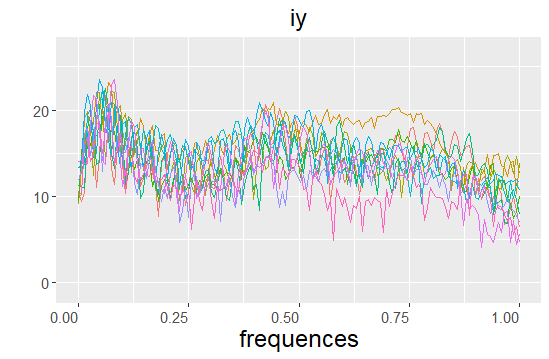}
\includegraphics[width=6.5cm, height=5cm]{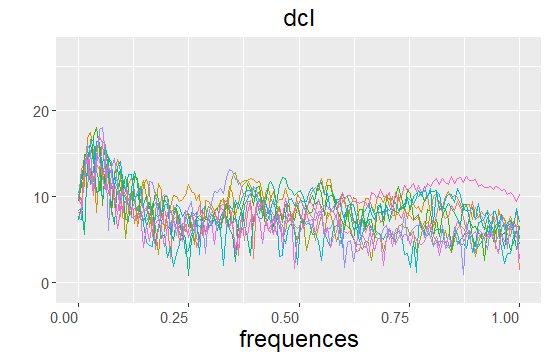}
\end{center}
\caption{A sample of 10 log-periodograms per class }
\label{FIG:data}
\end{figure}

This benchmark data example  was extracted from the TIMIT database 
(\url{https://catalog.ldc.upenn.edu/LDC93s1}), which is a widely used
resource for research in speech recognition and functional data classification. The data set we used was
constructed by selecting four phonemes for classification based on digitized
speech from this database.  From each speech
frame, a log-periodogram transformation is applied so as to cast the speech data
in a form suitable for speech recognition. The five phonemes in this data
set are transcribed as follows: ``sh'' as in ``she'', ``dcl'' as in
``dark'', ``iy'' as the vowel in ``she'', ``aa'' as the vowel in ``dark'',
and ``ao'' as the first vowel in ``water''. For illustration purpose, we
focus on the ``aa'', ``ao'', ``iy'' and ``dcl''  phoneme classes. Each speech frame is represented by $n=400$ samples at a 16-kHz sampling rate;   the first $M=150$ frequencies from each subject are retained.  Figure \ref{FIG:data} displays 10 log-periodograms for each class phoneme.

We randomly select training sample size $n_1=n_2=300$ to train the three classifiers  and the rest of $100$ samples remained as the test samples.
Network parameters were selected based on training data using Steps 1-3 in Section \ref{sec:FDNN}. Table \ref{TAB:Speech}  reports the mean percentage
(averaged over the $100$ repetitions) of misclassified test curves.  Overall,   FDNN   outperformed QD and NB   in all the four classification tasks. As depicted in the Figure 1, ``aa'' and ``ao'' phoneme class trajectories looks extremely similar to each other, the misclassification rate  is fairly larger than other classification results. The proposed classifiers   FDNN  still provides smallest risks and smaller standard errors compared with QD and NB classifiers. For ``ao'' vs``iy'', the misclassification rate  of   FDNN is less than one third of that of QD; For ``ao'' vs``dcl'', the misclassification rate of   FDNN is nearly half of that of NB.   


 \begin{table} \centering
   \caption{Misclassification rates ($\%$) with standard errors in brackets for Speech Recognition data.}
   \label{TAB:Speech}
 \begin{tabular}{@{\extracolsep{0.1pt}} crrrr}
 \hline
 \hline
 Classes &   \multicolumn{1}{c}{FDNN} & \multicolumn{1}{c}{QD} & \multicolumn{1}{c}{NB}\\
 \hline
 ``aa'' vs ``ao''  &  20.744(0.016)& 25.402(0.026) & 25.378(0.021)    \\ \hline
 ``aa'' vs ``iy''  &  0.193(0.002) & 0.288(0.005) &  0.273(0.006)   \\ \hline
 ``ao'' vs ``iy''  &    0.183(0.004) & 0.578(0.005) &  0.232(0.005)   \\ \hline
 ``ao'' vs ``dcl'' &    0.229(0.002) & 0.391(0.005) &  0.472(0.006)   \\ \hline
 \end{tabular}
 \end{table}

\subsection{ADNI database}\label{SEC:Real_2D}
The dataset used in the preparation of this article were obtained from the ADNI database (\url{adni.loni.usc.edu}).
The ADNI is a longitudinal multicenter study designed to develop clinical, imaging, genetic, and biochemical biomarkers for the early detection and tracking of AD.
From this database,  we collect PET data from $79$ patients in AD group, and $45$ patients in EMCI group. This  PET  dataset has been  spatially normalized and post-processed. These AD patients have three to six times doctor visits and we select the PET scans obtained in the third visits. People in EMCI group only have the second visit, and we select the PET scans obtained in the second visits. For AD group, patients' age ranges from $59$ to $88$ and average age is $76.49$, and there are $33$ females and $46$ males among these $79$ subjects. For EMCI group, patients' age ranges from $57$ to $89$ and average age is $72.33$, and there are $26$ females and $19$ males among these $45$ subjects.  All scans were reoriented into $79\times 95 \times 68$ voxels, which means each patient has $68$ sliced 2D images with  $79\times 95$ pixels.  For 2D case, it means each subject has $N=79\times 95=7,505$ observed pixels for each selected image slice.   For 3D case, the observed number of voxels for each patient's brain sample is $N=79\times 95 \times 68$.

\begin{figure}
\begin{center}
\hspace{.8cm}\textbf{AD} \hspace{5cm}\textbf{EMCI}\\
\hspace{.2cm}$5$-th
$\begin{array}{l}
\includegraphics[width = 0.35\textwidth]{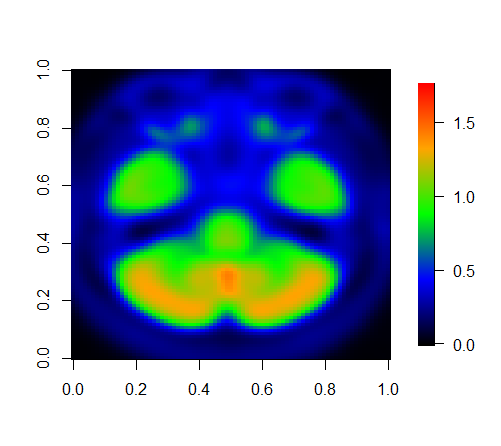} 
\hspace{2mm}
 \includegraphics[width = 0.35\textwidth]{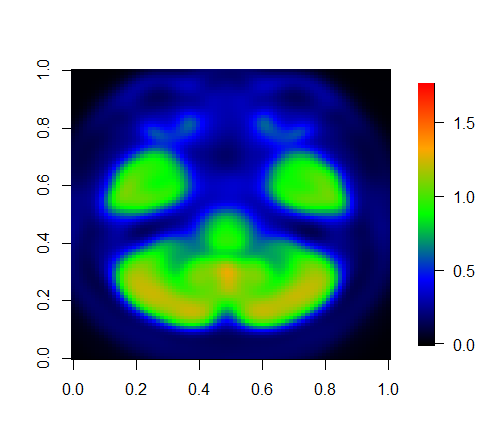} \\
\end{array}$\\
\vspace{-.6cm}
$10$-th
$\begin{array}{l}
\includegraphics[width = 0.35\textwidth]{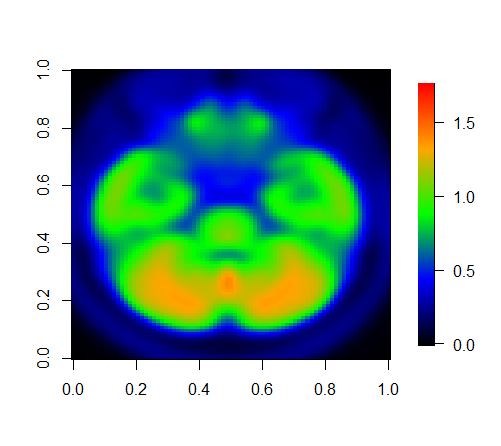} 
\hspace{2mm}
\includegraphics[width = 0.35\textwidth]{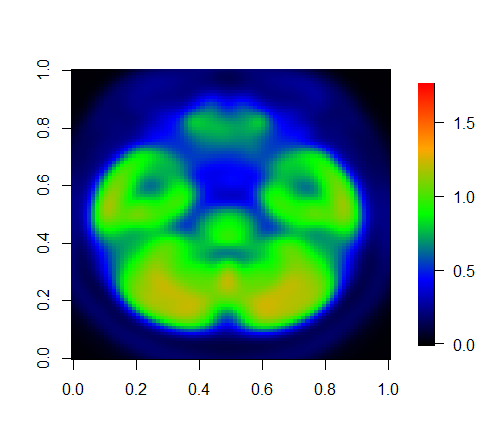} \\
\end{array}$\\
\vspace{-.6cm}
$15$-th
$\begin{array}{l}
\includegraphics[width = 0.35\textwidth]{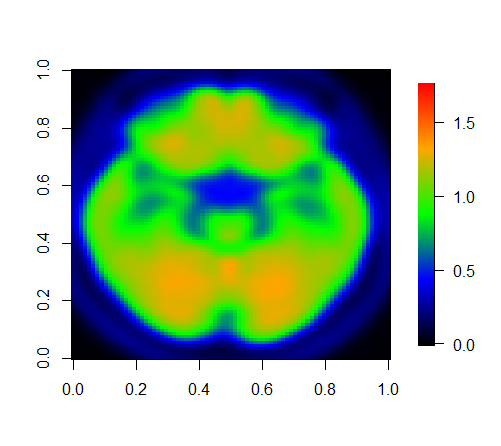} 
\hspace{1.mm}
 \includegraphics[width = 0.35\textwidth]{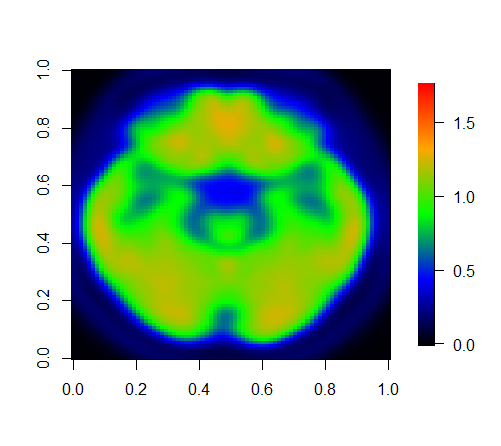} \\
\end{array}$\\
\vspace{-.6cm}
$20$-th
$\begin{array}{l}
\includegraphics[width = 0.35\textwidth]{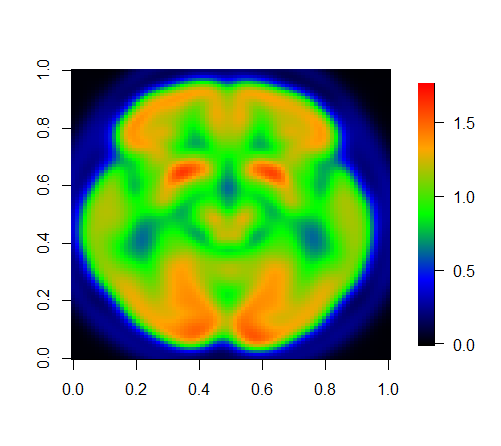} 
\hspace{2mm}
\includegraphics[width = 0.35\textwidth]{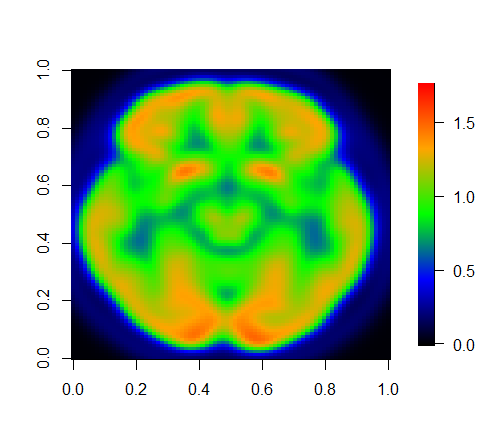} \\
\end{array}$\\
\vspace{-.6cm}
$25$-th
$\begin{array}{l}
\includegraphics[width = 0.35\textwidth]{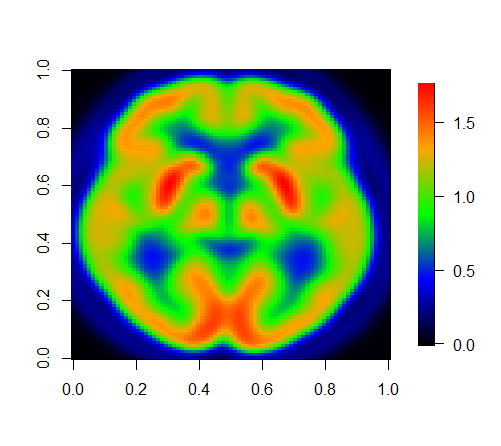} 
\hspace{1.mm}
 \includegraphics[width = 0.35\textwidth]{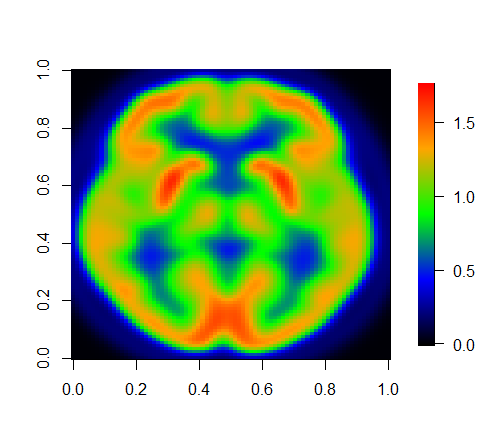} \\
\end{array}$\\
\vspace{-.6cm}
\caption{\label{fig:ADNI_2d} Averaged images of the $5$-th, the $10$-th, the $15$-th, the $20$-th and the $25$-th slices of EMCI (left column) group and AD  group (right column).} 
\end{center}
\end{figure}

\begin{figure}
    \centering
    \includegraphics[width = 0.70\textwidth]{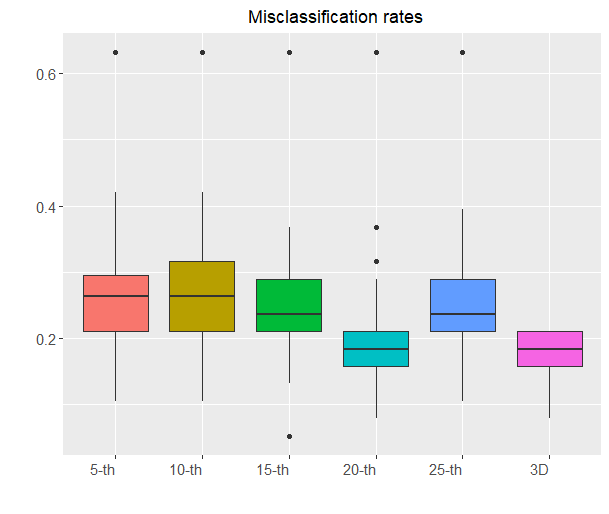} 
    \caption{\label{fig:ADNI_2d_boxplot}Grouped boxplot of misclassification rates for the $5$-th, the $10$-th, the $15$-th, the $20$-th , the $25$-th slices and $3$D data of the first $25$ slices  between EMCI and AD groups.}
\end{figure}

It is well known that Alzheimer's disease destroys neurons and their connections in hippocampus, the entorhinal cortex, and the cerebral cortex. These parts are corresponding to the first $25$ slices. Therefore, for our $2$D case study,  we specifically select the $5$-th, $10$-th, $15$-th, $20$-th and $25$-th slices from $68$ slices for each patient. We aim to conduct classification based on the information of those slices respectively; see \cite{MG:2011}. Figure \ref{fig:ADNI_2d} shows the averaged $2$D images for two groups at each slice.  For $3$D case, we  focus on the total $25$ slices, so the $3$D data is observed on $79\times 95 \times 25$ points. Figure \ref{fig:ADNI_2d_boxplot} demonstrates the misclassification rates for both $2$D and $3$D brain imaging data. There are several interesting finds. First,   given a single slice  $2$D imaging data, the misclassification rates tend  to be larger than using total $25$ slices data ($3$D data). It indicates that   $3$D data contains more helpful information to decrease the misclassification risk. 
Second,   the $20$-th slice provides the lowest one among all $2$D data. It is a promising finding for neurologists, as this smallest risk indicates this particular slice presents useful information to distinguish the EMCI and AD groups. Further medical checkups are meaningful for this special location in the brain.

\section{Conclusion}\label{SEC:discussion}
 We propose a new FDNN classifier for classifying non-Gaussian complex function data. 
Our contributions are twofold. First,
we establish sharp convergence rates for MEMR when data are of functional type, and the result can be applied to a large scope of functional data with complex density functions. The proposed FDNN is able to attain the sharp rate.
Second, our FDNN classifier is able to handle various 1D or multi-dimensional complex functional data. As demonstrated through extensive simulated
and real-data examples, the proposed FDNN classifier has outstanding performances in both Gaussian and non-Gaussian settings. 


\section*{Acknowledgement}
Wang's and  Cao's  research was partially supported by  NSF award DMS 1736470.   Cao's  research was also partially supported by    Simons Foundation under Grant \#849413.
Shang's research was supported in part by NSF DMS 1764280 and 1821157.

Data used in preparation of this article were obtained from the Alzheimers Disease Neuroimaging Initiative (ADNI) database (\url{adni.loni.usc.edu}). As such, the investigators within the ADNI contributed to the design and implementation of
ADNI and/or provided data but did not participate in analysis or writing of this report.
A complete listing of ADNI investigators can be found at: \url{http://adni.loni.usc.edu/wp-content/uploads/how_to_apply/ADNI_Acknowledgement_List.pdf}.

\bibliographystyle{plain}
\bibliography{Ref}  

\begin{thebibliography}{10}

\bibitem{adrover2004globally}
Jorge Adrover, Matias Salibian-Barrera, and Ruben Zamar.
\newblock Globally robust inference for the location and simple linear
  regression models.
\newblock {\em Journal of Statistical Planning and Inference}, 119(2):353--375,
  2004.

\bibitem{Audibert:Tsybakov:07}
Jean-Yves Audibert and Alexandre~B. Tsybakov.
\newblock Fast learning rates for plug-in classifiers.
\newblock {\em The Annals of Statistics}, 35:608--633, 2007.

\bibitem{Bauer:Kohler:19}
B.~Bauer and M.~Kohler.
\newblock On deep learning as a remedy for the curse of dimensionality in
  nonparametric regression.
\newblock {\em The Annals of Statistics}, 47:2261--2285, 2019.

\bibitem{Berrendero:etal:18}
J.~R. Berrendero, A.~Cuevas, and J.~L. Torrecilla.
\newblock On the use of reproducing kernel hilbert spaces in functional
  classification.
\newblock {\em Journal of the American Statistical Association},
  113(523):1210–1218, 2018.

\bibitem{bos2021arxiv}
Thijs Bos and Johannes Schmidt-Hieber.
\newblock Convergence rates of deep relu networks for multiclass
  classification.
\newblock {\em arXiv:2108.00969}, 2021.

\bibitem{Cai:Zhang:19b}
T.~Tony Cai and Linjun Zhang.
\newblock A convex optimization approach to high-dimensional sparse quadratic
  discriminant analysis.
\newblock {\em arXiv:1912.02872}, 2019.

\bibitem{Cai:Zhang:19}
T.~Tony Cai and Linjun Zhang.
\newblock High dimensional linear discriminant analysis: optimality, adaptive
  algorithm and missing data.
\newblock {\em Journal of the Royal Statistical Society. Series B. Statistical
  Methodology}, 81(4):675--705, 2019.

\bibitem{Chamroukhi:12}
F.~Chamroukhi and H.~Glotin.
\newblock Mixture model-based functional discriminant analysis for curve
  classification.
\newblock {\em Proceedings of the International Joint Conference on Neural
  Networks (IJCNN)}, pages 1--8, 2012.

\bibitem{Dai:etal:17}
Xiongtao Dai, Hans-Georg M\"{u}ller, and Fang Yao.
\newblock Optimal {B}ayes classifiers for functional data and density ratios.
\newblock {\em Biometrika}, 104(3):545--560, 2017.

\bibitem{Delaigle:Hall:12}
A.~Delaigle and P.~Hall.
\newblock Achieving near-perfect classification for functional data.
\newblock {\em Journal of the Royal Statistical Society, Series~B},
  74:267--286, 2012.

\bibitem{Delaigle:etal:12}
A.~Delaigle, P.~Hall, and N.~Bathia.
\newblock Componentwise classification and clustering of functional data.
\newblock {\em Biometrika}, 99(2):299--313, 2012.

\bibitem{Delaigle:Hall:13}
Aurore Delaigle and Peter Hall.
\newblock Classification using censored functional data.
\newblock {\em Journal of the American Statistical Association},
  108(504):1269--1283, 2013.

\bibitem{Farnia:Tse:16}
Farzan Farnia and David Tse.
\newblock A minimax approach to supervised learning.
\newblock In {\em Proceedings of the 30th International Conference on Neural
  Information Processing Systems}, NIPS'16, page 4240–4248, Red Hook, NY,
  USA, 2016.

\bibitem{Galeano:etal:15}
Pedro Galeano, Esdras Joseph, and Rosa~E. Lillo.
\newblock The {M}ahalanobis distance for functional data with applications to
  classification.
\newblock {\em Technometrics}, 57(2):281--291, 2015.

\bibitem{gam1990}
Travor~J. Hastie and Robert~J. Tibshirani.
\newblock {\em Generalized Additive Models}.
\newblock Chapman \& Hall/CRC, 1990.

\bibitem{HsingEubank2015}
T.~Hsing and R.~Eubank.
\newblock {\em Theoretical foundations of functional data analysis, with an
  introduction to linear operators}.
\newblock Wiley Series in Probability and Statistics. John Wiley \& Sons, Ltd.,
  Chichester, 2015.

\bibitem{hu2020arxiv}
Tianyang Hu, Zuofeng Shang, and Guang Cheng.
\newblock Sharp rate of convergence for deep neural network classifiers under
  the teacher-student setting.
\newblock {\em arXiv:2001.06892}, 2020.

\bibitem{Kim:NN:2021}
Yongdai Kim, Ilsang Ohn, and Dongha Kim.
\newblock Fast convergence rates of deep neural networks for classification.
\newblock {\em Neural Networks}, 138:179--197, 2021.

\bibitem{Kokoszka:Reimherr:17}
P.~Kokoszka and M.~Reimherr.
\newblock {\em Introduction to functional data analysis}.
\newblock Texts in Statistical Science Series. CRC Press, Boca Raton, FL, 2017.

\bibitem{Lecue:08}
Guillaume Lecu\'{e}.
\newblock Classification with minimax fast rates for classes of {B}ayes rules
  with sparse representation.
\newblock {\em Electronic Journal of Statistics}, 2:741--773, 2008.

\bibitem{Leng:Muller:06}
X.~Leng and H.G. M\"{u}ller.
\newblock Classification using functional data analysis for temporal gene
  expression data.
\newblock {\em Bioinformatics}, 22:68–76, 2006.

\bibitem{shang2021arxiv}
Kexuan Li, Fangfang Wang, Ruiqi Liu, Fan Yang, and Zuofeng Shang.
\newblock Calibrating multi-dimensional complex ode from noisy data via deep
  neural networks.
\newblock {\em arXiv:2106.03591}, 2021.

\bibitem{lin2000aos}
Yi~Lin.
\newblock Tensor product space anova models.
\newblock {\em The Annals of Statistics}, 28:734 -- 755, 2000.

\bibitem{shang2021jmaa}
Ruiqi Liu, Ben Boukai, and Zuofeng Shang.
\newblock Optimal nonparametric inference via deep neural network.
\newblock {\em Journal of Mathematical Analysis and Applications}, 505:125561,
  2022.

\bibitem{Liu:etal:2021}
Ruiqi Liu, Zuofeng Shang, and Guang Cheng.
\newblock On deep instrumental variables estimate.
\newblock {\em arXiv:2004.14954}, 2021.

\bibitem{Mammen:etal:99}
Enno Mammen and Alexandre~B. Tsybakov.
\newblock Smooth discrimination analysis.
\newblock {\em The Annals of Statistics}, 27:1808--1829, 1999.

\bibitem{Mazuelas:etal:20}
Santiago Mazuelas, Andrea Zanoni, and Aritz Perez.
\newblock Minimax classification with 0-1 loss and performance guarantees.
\newblock {\em arXiv:2010.07964}, 2020.

\bibitem{MG:2011}
Yangling Mu and Fred~H. Gage.
\newblock Adult hippocampal neurogenesis and its role in alzheimer's disease.
\newblock {\em Mol Neurodegener.}, 2011.

\bibitem{Park:etal:20}
Juhyun Park, Jeongyoun Ahn, and Yongho Jeon.
\newblock Sparse functional linear discriminant analysis.
\newblock {\em arXiv:2012.06488}, 2020.

\bibitem{Petersen:Voigtlaender:2018}
Philipp Petersen and Felix Voigtlaender.
\newblock Optimal approximation of piecewise smooth functions using deep relu
  neural networks.
\newblock {\em Neural Networks}, 108:296--330, 2018.

\bibitem{Rossi:05}
Fabric Rossi, Delannay Nicolas, Brieuc Conan-Guez, and Michel Verleysen.
\newblock Representation of functional data in neural networks.
\newblock {\em Neurocomputing}, 64:183--210, 2005.

\bibitem{Schmidt:19}
J.~Schmidt-Hieber.
\newblock Nonparametric regression using deep neural networks with relu
  activation function.
\newblock {\em The Annals of Statistics}, 48(4):1875--1897, 2020.

\bibitem{Shin:08}
H.~Shin.
\newblock An extension of fisher’s discriminant analysis for stochastic
  processes.
\newblock {\em Journal of Multivariate Analysis}, 99:1191–--1216, 2008.

\bibitem{Song:etal:08}
J.~Song, W.~Deng, H.~Lee, and D.~Kwon.
\newblock Optimal classification for time-course gene expression data using
  functional data analysis.
\newblock {\em Biometrika}, 103(1):147--159, 2016.

\bibitem{Steinwart:Christmann:08}
Ingo Steinwart and Andreas Christmann.
\newblock {\em Support vector machines}.
\newblock Springer Science and Business Media, 2008.

\bibitem{Suzuki:18}
Taiji Suzuki.
\newblock Adaptivity of deep relu network for learning in besov and mixed
  smooth besov spaces: optimal rate and curse of dimensionality.
\newblock {\em arXiv preprint arXiv:1810.08033}, 2018.

\bibitem{Tsybakov:04}
Alexandre~B. Tsybakov.
\newblock Optimal aggregation of classifiers in statistical learning.
\newblock {\em The Annals of Statistics}, 32:135--166, 2004.

\bibitem{Tsybakov:09}
Alexandre~B. Tsybakov.
\newblock {\em Introduction to nonparametric estimation. Springer Series in
  Statistics}.
\newblock Springer, New York, 2009.

\bibitem{Wang:etal:16}
J.L. Wang, J.~M. Chiou, and H.~G. M\"{u}ller.
\newblock Functional data analysis.
\newblock {\em Annual Review of Statistics and Its Application}, 3:257--295,
  2016.

\bibitem{wang2021stat}
Shuoyang Wang, Guanqun Cao, and Zuofeng Shang.
\newblock Estimation of the mean function of functional data via deep neural
  networks.
\newblock {\em Stat}, e393, 2021.

\bibitem{optfunctional2021arxiv}
Shuoyang Wang, Zuofeng Shang, Guanqun Cao, and Jun~S. Liu.
\newblock Optimal classification for functional data.
\newblock {\em arXiv:2103.00569}, 2021.

\bibitem{Yarotsky:17}
Dmitry Yarotsky.
\newblock Error bounds for approximations with deep relu networks.
\newblock {\em Neural Network}, 94:103--114, 2021.

\end{thebibliography}
\end{document}